\documentclass[sigconf]{acmart}
\AtBeginDocument{%
  \providecommand\BibTeX{{%
    \normalfont B\kern-0.5em{\scshape i\kern-0.25em b}\kern-0.8em\TeX}}}
\setcopyright{acmcopyright}
\copyrightyear{2018}
\acmYear{2021}\copyrightyear{2021}
\acmConference[WiseML '21]{3rd ACM Workshop on Wireless Security and Machine Learning}{June 28--July 2, 2021}{Abu Dhabi, United Arab Emirates}
\acmBooktitle{3rd ACM Workshop on Wireless Security and Machine Learning (WiseML '21), June 28--July 2, 2021, Abu Dhabi, United Arab Emirates}
\acmPrice{15.00}
\acmDOI{10.1145/3468218.3469047}
\acmISBN{978-1-4503-8561-9/21/06}

\begin{document}

\title{Adversarial Classification of the Attacks on Smart Grids Using Game Theory and Deep Learning}

\author{Kian Hamedani}
\affiliation{%
  \institution{ECE Department of Virginia Tech,  Blacksburg, VA, USA. \\
  Marconi-Rosenblatt AI/ML Innovation Lab,} 
  \country{ANDRO Computational Solutions LLC, Rome, NY, USA}
}
\email{hkian@vt.edu}

\author{Lingjia Liu}
\affiliation{%
  \institution{ECE Department of Virginia Tech}
  \country{Blacksburg, VA, USA}
}
\email{ljliu@vt.edu}

\author{Jithin Jagannath}
\affiliation{%
  \institution{ Marconi-Rosenblatt AI/ML Innovation Lab,}
  \country{ANDRO Computational Solutions LLC, Rome, NY, USA}
}
\email{jjagannath@androcs.com}
\author{Yang (Cindy) Yi}
\affiliation{%
  \institution{ECE Department of Virginia Tech}
  \country{Blacksburg, VA, USA}
}
\email{yangyi8@vt.edu}
\begin{abstract}
Smart grids are vulnerable to cyber-attacks. This paper proposes a game-theoretic approach to evaluate the variations caused by an attacker on the power measurements. Adversaries can gain financial benefits through the manipulation of the meters of smart grids. On the other hand, there is a defender that tries to maintain the accuracy of the meters. A zero-sum game is used to model the interactions between the attacker and defender. In this paper, two different defenders are used and the effectiveness of each defender in different scenarios is evaluated. Multi-layer perceptrons (MLPs) and traditional state estimators are the two defenders that are studied in this paper. The utility of the defender is also investigated in adversary-aware and adversary-unaware situations. Our simulations suggest that the utility which is gained by the adversary drops significantly when the MLP is used as the defender.  It will be shown that the utility of the defender is variant in different scenarios, based on the defender that is being used. In the end, we will show that this zero-sum game does not yield a pure strategy, and the mixed strategy of the game is calculated.
\end{abstract}


\maketitle



\section{Introduction}
Smart grids are new infrastructure that integrates energy by many different technologies, such as telecommunication, the internet, and electronic devices in this era of Internet-of-Things (IoT). This convergence of different technologies brings up some opportunities and challenges as well. The main opportunity that smart grids provide is a bidirectional flow of electricity and information between power suppliers and customers, which will result in a more efficient distribution of power. However, due to the integration of different technologies, the smart grids are more vulnerable to cyber-attacks \cite{Ozay2016}. False data injection (FDI) attacks are known to be one of the most malicious cyber-attacks in smart grids\cite{Reiter2011}. Cyber security tries to maintain  reliable and secure communication between different components of the network, including communication networks and computer systems\cite{Liang2017rev}. As a result of this secure communication, supervisory control and data acquisition (SCADA) system can have a better estimation of the network state. State estimation is a critical process in a control system and due to this fact, the SCADA is usually a target for attackers. Injecting false data to the SCADA manipulates the state estimation and it can cause economic gains for the attacker\cite{Liang2017rev}.   

Game theory has shown to be a powerful approach to model and capture the complex interactions among the different players of electricity markets, which in some cases have a conflict of interests together\cite{Srinivasan2017}. During the last decades, game theory has been applied in many different fields, including economics, politics, and psychology. Due to the demand of having intelligent, autonomous, and flexible networks, in which devices can make rational decisions, game theory has also been applied in wireless communication networks and smart grids\cite{Srinivasan2017,Saad2009,wang2020leveraging,li2020risk}. Different devices or software can be the players in the network security domain. These players make their decisions independently and can be cooperative, non-cooperative, or even malicious towards each other\cite{Manshaei2013}. As a result of this interaction among the players, the rational strategic decisions that they affect the network security and can be modeled as a game\cite{Manshaei2013}. However, it is essential to consider the role of bad data detectors (BDD) while analyzing the strategic decisions and interactions among the players. The game theory can only study the interactions between the attackers and defenders, while the machine learning methods are being vastly used as the BDD. Therefore, it is necessary to incorporate the machine learning-based BDDs in the game-theoretical formulation of the smart grids. 

Machine learning-based approaches have also been widely studied for several applications in the realm of IoT \cite{JagannathAdHoc2019}. Similarly, machine learning has been applied for FDI detection in smart grids \cite{Ozay2016}. Artificial neural networks, support vector machines (SVM), and k-nearest neighbor (KNN) are few examples of different machine learning algorithms that have been used for this purpose \cite{Ozay2016}. Machine learning-based approaches have shown better performance than the traditional state vector estimation (SVE) in detecting the FDI in smart grids. Esmalifalak \textit{et al} \cite{Ese2014} have applied dimension reduction for mapping the data collected from the network and used both supervised and unsupervised machine learning algorithms to detect the stealth data injections in smart grids. In \cite{Yan2014}, the performance of SVM, and KNN in detecting the false data in an IEEE-30 bus system under balanced and imbalanced data scenarios are studied. Mohammadpourfard \textit{et al} \cite{Mohammadpourfard2017} proposed an unsupervised anomaly detection method in smart grids which considers the effect of wind power generation and topology configuration. Zhao \textit{et al} \cite{Scala2017} have proposed a method based on short-term state forecasting which considers a temporal correlation among measurements. Moslemi \textit{et al} \cite{Moslemi2017}, proposed an approach based on maximum likelihood (ML) which is decentralized, and the near chordal sparsity of smart grids is considered to detect the FDI. Esmailfalak \textit{et al} \cite{esm2013game} proposed a game theory-based approach to study the utility of the attacker and defender, while the defender is using a traditional state estimator as the BDD.

In this paper, we use a zero-sum game theory approach to quantify the utility gain of the attacker. We will show that different choices of the defender can significantly affect the gain of the attacker. In one scenario, the traditional state estimator, and in the other scenario an MLP is selected as the defender. We assume that the attacker and the defender are not able to attack and defend all the meters, andcompetition exists between attacker and defender to maximize and minimize the variations caused in smart meters due to false data injection respectively. The attacker can gain higher financial utility if the false variations of the measurements read by the meters are maximized. We will demonstrate that when the defender is chosen to be based on an MLP, the utility that the attacker can gain is much less than the case the defender is a traditional state estimator. The utility of the defender is also studied in different scenarios.  The utility of the defender is defined as the number of the compromised measurements that the defender can accurately detect. Adversary-aware and adversary-unaware are the two main scenarios that are used to study the utility of the defender. In the adversary-aware situation, the attacker is aware of the existence of the defender and tries to maximize its attack utility concerning the parameters of the defender. On the other hand, in an adversary-unaware scenario, the attacker is unaware that there exists a  defender and it optimizes its attack regardless of the defender parameters. We will show that the utility of the defender is different in these two situations. So far in the literature, only state estimators have been used as the BDD in the game-theoretic formulations of smart grids security and to the best of our knowledge, this is the first time that MLPs have been used as the BDD in the game-theoretic approach of studying smart grids security. 

The organization of this paper is as follows: Section II represents the problem formulation. In Section III, the simulation results are discussed. Section IV represents the conclusion.

\section{Problem Formulation}
\subsection{System Model}
In direct current (DC) power flow transmission, a linear equation is used to approximate the power flow\cite{esm2013game}:

\begin{equation}
P_{ij}=\frac{x_{i}-x_{j}}{R_{ij}},
\label{eq1}
\end{equation}
where $P_{ij}$ is the power flowing between bus $i$ and $j$, $x$ is the bus voltage phase angle, and $R_{ij}$ is the reactance of transmission line between buses $i$ and $j$\cite{esm2013game}. The control center observes the measurement vector $\textbf{z}$ \cite{ham2018}:

\begin{equation}
\textbf{z}=\textbf{H}\textbf{x} + \textbf{e},
\label{eq2}
\end{equation}
where $\textbf{z}$ is the measurements vector, $\textbf{H}$ is the Jacobian matrix, $\textbf{x}$ is the state vector, and $\textbf{e}$ is the environment noise. The attacker can compromise the state estimation by injecting false data, $\textbf{a}$. The attack formulation can be expressed as follows:

\begin{equation}
\textbf{z}=\textbf{Hx} + \textbf{a} + \textbf{e}.
\label{eq3}
\end{equation}

The estimated state vector can be calculated as:

\begin{equation}
\hat{\textbf{x}}=(\textbf{H}^{T}\wedge^{-1}\textbf{H})^{-1}\textbf{H}^{T}\wedge^{-1}\textbf{z} = \textbf{Mz},
\label{eq4}
\end{equation}
where $\textbf{M}=(\textbf{H}^{T}\wedge^{-1}\textbf{H})^{-1}\textbf{H}^{T}\wedge^{-1}$, and $\wedge$ is the noise covariance matrix.
\subsection{Attack Against State Estimation}
The real time pricing can be affected if the adversaries can successfully launch attacks on the results of the real time state estimation. The main goal of the attacker is to perform the attack in a way that it will not be detected by the BDD. The first thing that the attacker needs to know, is the group of measurements that can increase or decrease the congestion in the transmission line after the false data is injected. It can be seen in Eq.~\ref{eq1} that any changes in the state (voltage phase angle) of any bus can change the power measurements of transmission lines. Combining Eq.~\ref{eq1} and ~\ref{eq4},  gives us the estimated power as follows\cite{esm2013game}:
\begin{equation}
  \begin{aligned}
  \hat{\textbf{P}_{ij}}=\frac{\hat{\textbf{x}_{i}}-\hat{\textbf{x}_{j}}}{\textbf{R}_{ij}}=\frac{(\textbf{M}_{i}-\textbf{M}_{j})^{T}}{\textbf{R}_{ij}}\textbf{z}\\
         =    \textbf{G}^{T}\textbf{z}= \textbf{G}_{+}^{T}\textbf{z}_{+}+\textbf{G}_{-}^{T}\textbf{z}_{-}  ,
  \end{aligned}
  \label{eq5}
\end{equation}
where $\textbf{G}^{T}=\frac{(\textbf{M}_{i}-\textbf{M}_{j})^{T}}{\textbf{R}_{ij}}$. The group of positive and negative measurements where increasing or decreasing their values by adding a false data, $\textbf{z}_{a} > 0$, leads to increasing or decreasing the congestion in transmission line are shown with $\textbf{z}_{+}$ and $\textbf{z}_{-}$, respectively. In this paper we assume that $\textbf{z}_{+}$ and $\textbf{z}_{-}$ belong to groups $K$ and $L$, respectively. It is also assumed in this paper that the attacker knows $\textbf{H}$, and as a result, it can make a distinction between groups $K$ and $L$. Adversaries try to perform their attacks in a way that cannot be detected by the BDD. In the case that BDD is chosen as the traditional state estimator, the residual value between the actual measurements and the estimated values, $\rho$, has to be less than a certain threshold value, $\zeta$, that the attack cannot be detected by the BDD. The residual value is calculated as follows:
\begin{equation}
  \begin{aligned}
\rho=\textbf{z}-\textbf{H}\hat{\textbf{x}}=\textbf{z}_{0}+\textbf{z}_{a}-\textbf{H}(\textbf{M}\textbf{z}_{0}+\textbf{M}\textbf{z}_{a})\\
=\textbf{z}_{0}-\textbf{H}\textbf{M}\textbf{z}_{0} + \textbf{z}_{a}-\textbf{H}\textbf{M}\textbf{z}_{a}= \rho_{0}+\rho_{a}
 \end{aligned}
 \label{eq6}
\end{equation}
where $\textbf{z}_{0}$ is the safe measurement vector, $\textbf{z}_{a}$ is the measurement vector which is compromised by the false data injected by the adversary, $\rho_{0}=\textbf{z}_{0}-\textbf{HM}\textbf{z}_{0}$, and $\rho_{a}=\textbf{z}_{a}-\textbf{HM}\textbf{z}_{a}$. It can be seen in Eq.~\ref{eq6} that $\rho_{a}$ corresponds to the residual value caused by the false data injected by the adversary\cite{esm2013game}. In order to perform the attack such that it cannot be detected, $\rho_{a}$ has to be smaller than a certain threshold value,
\begin{equation}
\rho_{a}=||(\textbf{I}-\textbf{HM})\textbf{z}_{a}|| \leq \zeta.
 \label{eq7}
\end{equation}
This threshold value, $\zeta$, is a design parameter that has to be specified by the attacker. The smaller values of $\zeta$ make it more challenging for the BDD to detect the attack. However, the small values of $\zeta$ will provide less gain for the attacker, because the attacker faces more strict constraints to manipulate the measurements. Adversaries have to solve an optimization problem to be able to perform the maximum manipulation on the measurements without being detected by the BDD. This optimization problem is expressed in Eq.~\ref{eq8}\cite{esm2013game}.
\begin{equation}
  \begin{aligned}
&\underset{\textbf{z}_{a}}{\mathrm{max}}\;\; \sum_{i\epsilon K} \textbf{z}_{a}(i)-\sum_{j\epsilon L} \textbf{z}_{a}(j)\\
&s.t.
    \begin{cases}
      ||(\textbf{I}-\textbf{HM})\textbf{z}_{a}|| \leq \zeta \\
      \textbf{z}_{a}(k)=0\;\; k \; \epsilon \; \{NK\},
    \end{cases}
  \end{aligned}
   \label{eq8}
\end{equation}
where $\{NK\}$ is the set of measurements that the attacker is not able to compromise.  The utility of the attacker corresponds to values of $\textbf{z}_{a}$ that are achieved from Eq. 8. It means that for the higher values of $\textbf{z}_{a}$, the measurements are manipulated more significantly and the attacker can gain higher utility. Online pricing of the power depends on the estimated power of the smart meters. Therefore, the attacker can gain some financial utility through injecting false data and changing the actual estimations of the measurements. The utility of the attacker can be written as follows:
\begin{equation}
\Delta\hat{\textbf{P}_{ij}}=\frac{(\textbf{M}_{i}-\textbf{M}_{j})^{T}}{\textbf{R}_{ij}}\textbf{z}_{a}.
 \label{eq9}
\end{equation}
\subsection{Attack Against Artificial Neural Networks}
\begin{figure}[htbp]
	\centering
	\includegraphics[width=0.47\textwidth,height=0.2\textheight]{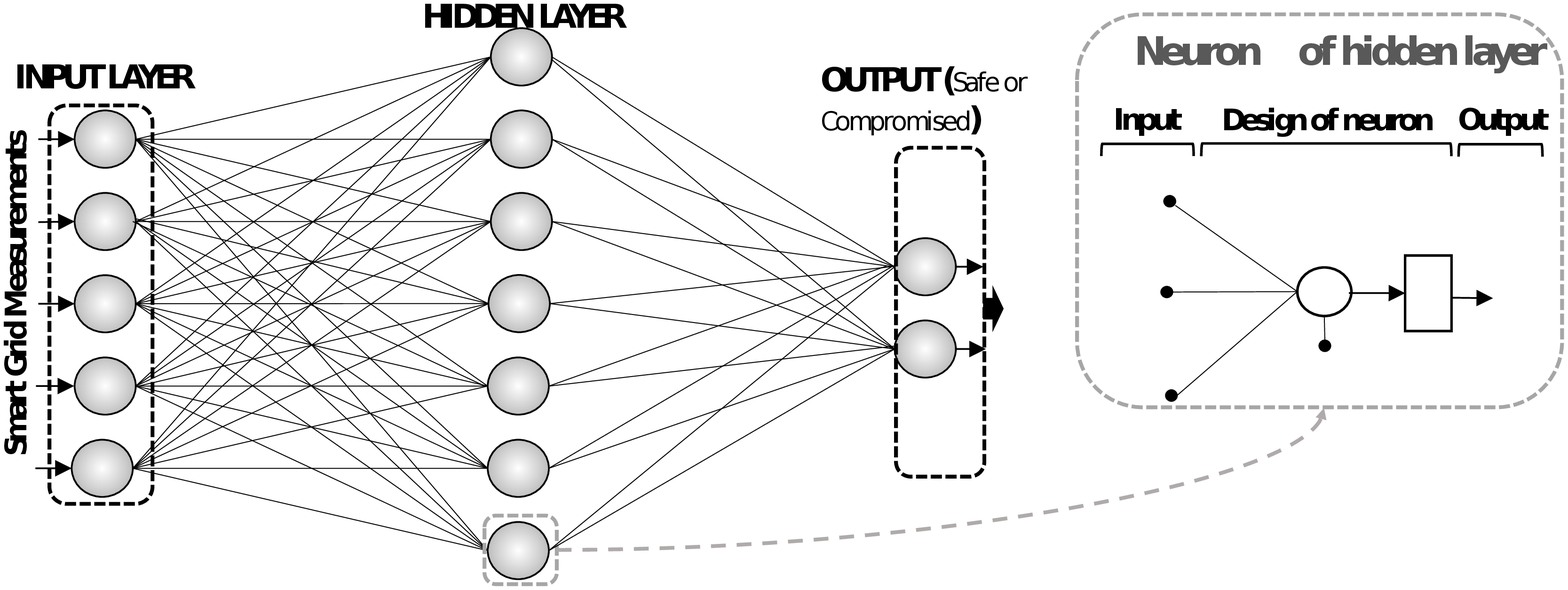}\par\vspace{-0.2cm}\caption{Structure of a MLP with one hidden layer.}
	\label{Fig1}
\end{figure}

In Section II B, we assumed that the BDD is a closed- form state estimator. However, it is possible to use other tools as the BDD. In this section, we propose the idea to use deep learning (MLP) instead of a state estimator. As it was mentioned in Sections II A and B, the attacker injects the false data in a way that cannot be detected by the BDD. In the case of using MLP as the BDD, the attacker has to perform its attack in a way that cannot be detected by the MLP. It means that the attacker needs to solve the optimization problem expressed in Eq. 8. However, the constraints of that optimization problem have to be modified with respect to the parameters of the MLP. In Section II B, it was assumed that the attacker knows $\textbf{H}$, in the case that MLP is used as the BDD we will assume that the attacker knows the parameters of the MLP. In Fig.~\ref{Fig1}, the structure of a one-layer MLP is depicted. There are two sets of weights in this network, $w_{ij}$, which connects the input to the first hidden layer, and $y_{ij}$ that connects the outcome of the hidden layer to the output layer. There are also two activation functions $f$ and $f1$. In this scenario, an MLP has to be trained with two sets of data, compromised data and safe data. For each set of data, a label is assigned to train the MLP, the label $1$ is assigned for the compromised measurements, and $0$ for safe data. The adversary has to perform its attack somehow that when the compromised data is tested by the MLP, the output of MLP will be closer to zero than one. In this paper, we will use an MLP with two hidden layers and 10000 total samples are used for training the MLP, half of the samples are compromised measurements and the other half are safe measurements. We assume that when the output of MLP is larger than $0.5$, the measurement will be classified as the compromised data, and when it is less than $0.5$, it will be considered as a safe measurement. Therefore,  Eq. 8 is expressed as follows when the BDD is an MLP:
\begin{equation}
  \begin{aligned}
&\underset{\textbf{z}_{a}}{\mathrm{max}}\;\; \sum_{i\epsilon K} \textbf{z}_{a}(i)-\sum_{j\epsilon L} \textbf{z}_{a}(j)\;\;\;\;\;\;\;\;\;\;\;\;\\
&s.t.
    \begin{cases}
      f1(y_{11}f(z(1)w_{11}+...+z(N)w_{N1})\\+y_{21}f(z(1)w_{12}+...+z(N)w_{N2})\\+...\\+y_{M1}f(z(1)w_{1N}+...+z(N)w_{NM})) \leq 0.5 \\
      z_{a}(k)=0\;\; k \; \epsilon \; \{NK\},
    \end{cases}
  \end{aligned}
   \label{eq10}
\end{equation}
where $z=z_{0}+z_{a}$, $N$ is the number of measurements, and $M$ is the number of neurons in the hidden layer. This formulation can be extended to any number of neurons and hidden layers.
\vspace{-0.5cm}
\subsection{Zero-Sum Game Between Two Players}
In a two-person zero-sum game there are two players that each side tries to increase its gain by decreasing the other side's gain. The adversary and the defender can also be considered as two players of a game. Each player has a set of strategies, which in our case is defined as the measurements that can be attacked and defended. Based on the strategy set, for any action taken by the attacker or defender, a utility value is attributed. In this paper, $\textbf{S}=\{s_{ij, i=1,...,r;j=1,...,c}\}$ is denoted as the game matrix. In the game matrix, the set of actions and the corresponding utilities which are taken by each player is shown. Here the utility of the attacker, $U_{a}$, corresponds to $\Delta\hat{P_{ij}}$ and the utility of defender, $U_{D}$, is equal to -$\Delta\hat{P_{ij}}$. The saddle pint or the pure strategy of this game is defined as $s_{i^{*}j^{*}}$, and it exists if and only if, $min(\underset{row}{\mathrm{max}})$ of $\mathbf{S}$=$max(\underset{col}{\mathrm{min}})$ of $\mathbf{S}$ where $row$ and $col$ correspond to the rows and columns of the $\textbf{S}$. In the games that a pure strategy does not exist, a mixed strategy of the game has to be calculated. The mixed strategy tells us about the probability of taking each action by the players\cite{esm2013game}. The frequency of choosing the rows and columns of the game by the defender and the attacker will converge to the probability distribution of the defender and attacker respectively. To calculate the mixed strategy of a zero-sum game, the defender tries to minimize the average value of the outcome of the game which is defined as below:
\begin{equation}
   min \; C(q,u)= \sum_{i=1}^{r}\sum_{j=1}^{c}q_{i}^{'}s_{ij}u_{j},
    \label{eq11}
\end{equation}
where $q$ and $u$ are the probability distributions of the strategies taken by the defender and attacker respectively. The defender's mixed strategy is calculated via solving the following optimization problem which is a linear programming problem:
\begin{equation}
\begin{aligned}
&\underset{\tilde{q}}{\mathrm{max}}\;\tilde{q}^{'}\;1_{r}\;\;\;\;\;\;\;\;\;\;\;\;\;\;\; \\
&s.t.
\begin{cases}
S^{'}\tilde{q}\;\leq\;1_{c}\;\;\;\;\;\;\\
\tilde{q}\;\geq\; 0,
     
\end{cases}
\end{aligned}
 \label{eq12}
\end{equation}
where $\tilde{q}=\dfrac{q}{h_{1}(q)}$, and  $h_{1}(q)$ is defined as follows:
\begin{equation}
h_{1}(q)=\underset{U}{\mathrm{max}}\;q^{'}S\;u\;\geq\;q^{'}S\;u\;\;\;\; \forall\;u\;in\;U.
 \label{eq13}
\end{equation}
In Eq.~\ref{eq13}, $U$ is defined as the probability distribution of the strategies chosen by the attacker and is defined as below:
\begin{equation}
U = \left\lbrace u \in R^{c}: u \geq 0,\sum_{j=1}^{c}u_{j}=1 \right\rbrace
 \label{eq14}
\end{equation}
In order to calculate the mixed strategy of the attacker another linear programming problem has to be solved which is expressed as:
\begin{equation}
\begin{aligned}
\underset{\tilde{u}}{\mathrm{min}}\;\tilde{u}^{'}\;1_{c}\;\;\;\;\;\;\;\;\;\;\;\;\;\;\; \\
s.t.\;\;\;\;\;\;\;\;\;\;\;\;\;\;\;\;\;\;\;\;\;\;\;\\
\begin{cases}
S\tilde{u}\;\geq\;1_{r}\;\;\;\;\;\;\\
\tilde{qu}\;\geq\; 0,
\end{cases}
\end{aligned}
 \label{eq15}
\end{equation}
where $\tilde{u}=\dfrac{u}{h_{2}(u)}$, $u_{i}^{'}$ is the transpose of $u$, and  $h_{2}(u)$ is defined as follows:
\begin{equation}
h_{2}(u)=\underset{Q}{\mathrm{min}}\;q^{'}S\;u\;\leq\;q^{'}S\;u\;\;\;\; \forall\;q\;in\;Q.
 \label{eq16}
\end{equation}

In Eq.~\ref{eq16}, $Q$ is defined as the probability distribution of the strategies chosen by the defender and is expressed as below:
\begin{equation}
Q = \left\lbrace q \in R^{r}: q \geq 0,\sum_{i=1}^{r}q_{i}=1 \right\rbrace
 \label{eq17}
\end{equation}

\section{Performance Analysis}
In this paper we use MATPOWER $5.1$ \cite{Zimmerman2011} to simulate a PJM-5 bus test system which gives us $\textbf{H}$, and $\textbf{x}$. To have a fair comparison between the MLP and the state estimator, the threshold value of the state estimator BDD, $\zeta$, has to be chosen somehow that the false alarm rate of the MLP and the state estimator be equal. In Table \ref{tab:table1} the $\zeta$ thresholds of the state estimator are shown when the attacker can perform its attack on one or two meters at the same time depending on which meters are being defended and which meters are being attacked in the zero-sum game. In the PJM-5 testbed, there are total of 11 smart meters. However, as it was mentioned in Section II we assume that the adversary is not able to perform an attack on all of them.
\begin{table}[!htbp]
  \begin{center}
    \caption{Threshold values of state estimator BDD in mega Watts(MW).}
    \label{tab:table1}
    \begin{tabular}{l|c} 
      \textbf{Compromised meter} & \textbf{$\zeta$}\\
      \hline
      $z_{1}$ & 8.54\\
      $z_{4}$ & 7.5 \\
      $z_{5}$ & 8.32 \\
      $z_{10}$& 8.3\\
      $z_{1}z_{4}$ & 10.89\\
      $z_{1}z_{5}$ & 12.65\\
      $z_{1}z_{10}$ & 12.59\\
      $z_{4}z_{5}$ & 10.48\\
      $z_{4}z_{10}$ & 10.33\\
      $z_{5}z_{10}$ & 9.48\\
    \end{tabular}
  \end{center}
\end{table}
Adversary performs its attack on the meters that are more likely to cause congestion on transmission lines. These meters are identified through DC optimal power flow (DCOPF) simulation and are shown in Table I. In Tables II and III, the utility of the attacker, in the zero-sum game between the attacker and the defender is shown. As it was mentioned in Section II D, in the zero-sum game the utility of the attacker, $U_{a}$, corresponds to $\Delta\hat{P_{ij}}$ and the utility of defender, $U_{D}$, is equal to -$\Delta\hat{P_{ij}}$. The columns of Tables II and III correspond to the meters that are attacked, and the rows correspond to the meters that are defended.

As it can be seen in Tables II and III, the utility of the attacker when the defender is using the MLP as the BDD is much lower than the case that the defender uses state estimator. The average utility of the attacker in the zero-sum game when the defender uses MLP is 0.87 MW and when it uses the state estimator, it is equal to 4.75 MW, which indicates that the MLP is a much better BDD for the defender since the attacker can gain much less utility.
\begin{table}[!htbp]
  \begin{center}
\centering
\caption{Utility of Attacker when BDD is state estimator(MW)}
  \begin{tabular}{|c|c| c| c| c| c|c|}
  \hline
 \hbox{Def}{Att}& z1z4 & z1z5 & z1z10 & z4z5 & z4z10 & z5z10 \\
 \hline
 z1z4 & 0 & 6.03 & 5.83 & 6.03 & 5.83 & 11.82 \\
 \hline
 z1z5 & 2.79 & 0 & 5.83 & 2.79 & 8.34 & 5.83\\
 \hline
 z1z10 & 2.79 & 6.03 & 0 & 8.6 & 2.79 & 6.03\\
 \hline
 z4z5 & 4.45 & 4.45 & 10.18 & 0 & 5.83 & 5.83\\
 \hline
 z4z10 & 4.45 & 10.44 & 4.45 & 6.03 & 0 & 6.03\\
 \hline
 z5z10 & 7.71 & 4.45 & 4.45 & 2.79 & 2.79 & 0\\
 \hline
  \end{tabular}
  \end{center}
\end{table}
\begin{table}[!htbp]
  \begin{center}

\centering
\caption{Utility of Attacker when BDD is MLP(MW)}
  \begin{tabular}{|c|c|c|c|c|c|c|}
  \hline
 \hbox{Def}{Att}& z1z4 & z1z5 & z1z10 & z4z5 & z4z10 & z5z10 \\
 \hline
 z1z4 & 0 & 0.77 & 1.5 & 0.77 & 1.5 & 2.33 \\
 \hline
 z1z5 & 0.86 & 0 & 1.5 & 0.86 & 1.62 & 1.5\\
 \hline
 z1z10 & 0.86 & 0.77 & 0 & 1.83 & 0.86 & 0.77\\
 \hline
 z4z5 & 0.28 & 0.28 & 1.28 & 0 & 1.5 & 1.5\\
 \hline
 z4z10 & 0.28 & 0.78 & 0.28 & 0.77 & 0 & 0.77\\
 \hline
 z5z10 & 0.77 & 0.28 & 0.28 & 0.86 & 0.86 & 0\\
 \hline
  \end{tabular}
  \end{center}
\end{table}\\
\vspace{-1cm}
\subsection{Adversary-aware VS Adversary-unaware classification}
\begin{figure}
	\centering
	\includegraphics[width=0.43\textwidth,height=0.18\textheight]{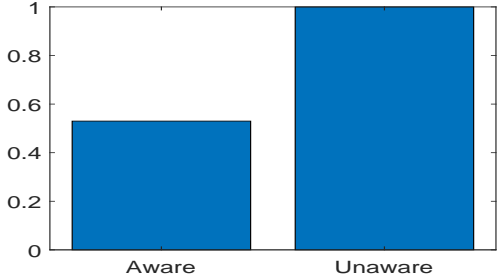}\par\vspace{-0.2cm}\caption{Detection probability of MLP.}
	\label{Fig2}
\end{figure}

\begin{figure}
	\centering
	\includegraphics[width=0.43\textwidth,height=0.18\textheight]{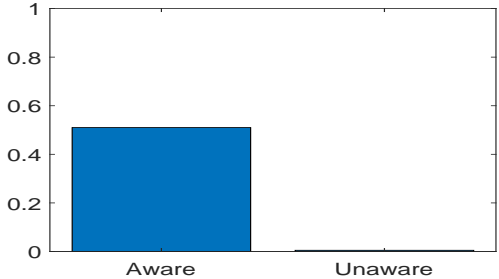}\par\vspace{-0.2cm}\caption{Detection probability of state estimator.}
	\label{Fig3}
	\vspace{-0.2in}
\end{figure}

 In adversarial attacks machine learning, the adversary tries to gain the maximum benefit through fooling the classifier. In this area, there exists a game or competition between the adversary and the classifier. Two main scenarios are considered in this situation. In the first scenario, it is assumed that the adversary is not aware of the presence of the BDD or the classifier, while in the second scenario it is assumed that the adversary is aware of that. The adversary-aware scenario requires the adversary to optimize its attack with respect to the parameters of the classifier, and the adversary-unaware requires the attacker to optimize its attack regardless of the classifier parameters\cite{Dalvi2004}. In this part, we first assume that the classifier is an MLP and then we assume that the classifier or the BDD is the state estimator. For both of the cases, the adversary-aware and unaware scenarios are implemented and the normalized utility of the classifiers are shown in Figs.~\ref{Fig2} \& ~\ref{Fig3}. The utility of the classifier is defined as the number of the compromised meters that it can detect.
 
As it can be seen in Figs.~\ref{Fig2} \& ~\ref{Fig3}, when the defender is using the MLP as the BDD, in the adversary-unaware case the utility or accuracy of the attack detection is higher compared with the adversary-aware case. However, when the defender is using the state estimator as the BDD, the opposite behavior is observed. 
The reason for this difference in the behavior is that when the BDD is MLP and the attacker optimizes its attacks with respect to the state estimator, the magnitude of the attack will be much higher which makes it very easy for the MLP to detect the attack. On the other hand, when the BDD is using the state estimator, and the attacker is optimizing its attack against the MLP the magnitude of the attack will be very low and the state estimator can't detect the attack. That is why in the adversary-unaware scenario, the utility of the classifier when MLP is being used is much higher than the case that the state estimator is being used. However, in the adversary-aware scenario, the utility of the classifier is almost equal in both cases, because the threshold value, $\zeta$, of the state estimator is chosen such that both the classifiers have the same value of false alarm rate.
\vspace{-0.25cm}
\subsection{Mixed Strategy of the Game}
\begin{figure}
\centering
	\includegraphics[width=0.4\textwidth,height=0.15\textheight]{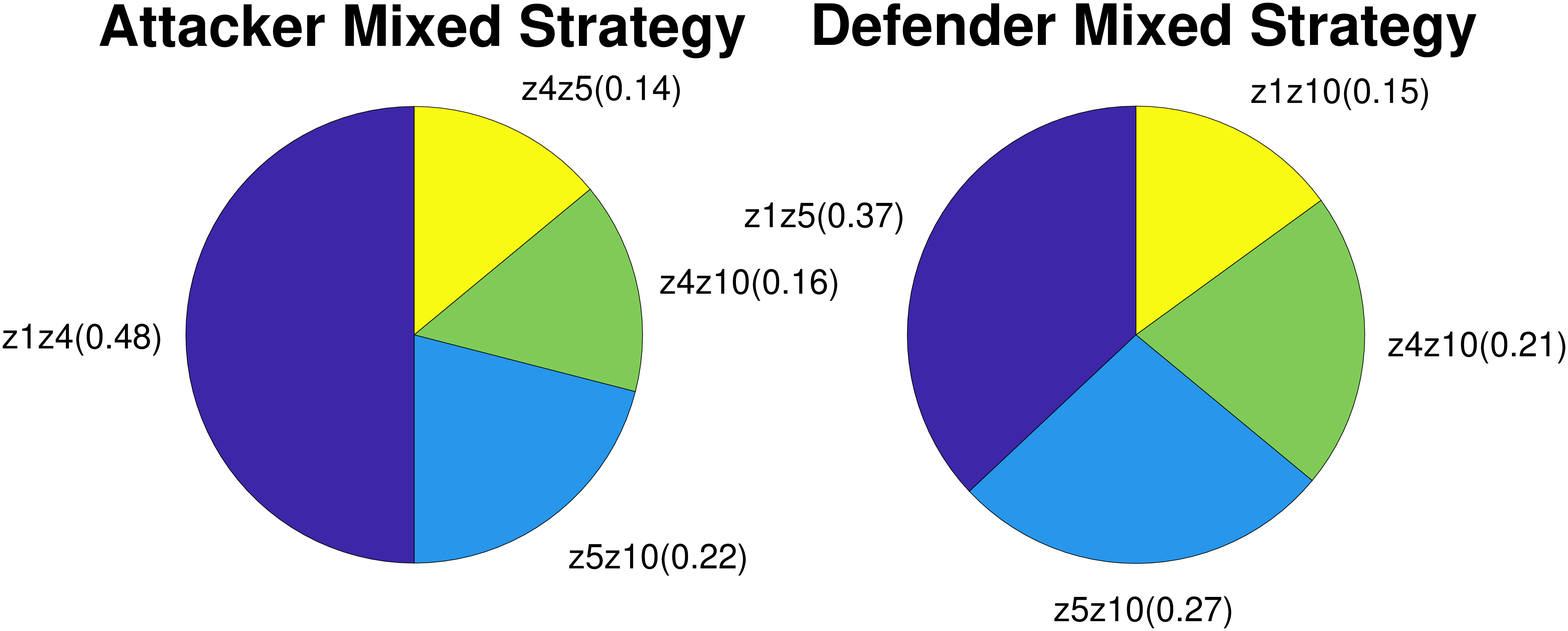}\par\vspace{-0.2cm}\caption{Mixed Strategy of Table II.}
	\label{Fig4}
\end{figure}

\begin{figure}
\centering
	\includegraphics[width=0.4\textwidth,height=0.15\textheight]{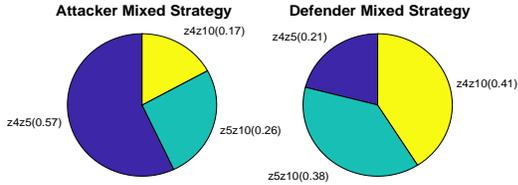}\par\vspace{-1cm}\caption{Mixed Strategy of Table III.}
	\label{Fig5}
\end{figure}

As it can be seen in Tables II and III, this game does not have a pure strategy because $min(\underset{row}{\mathrm{max})}$ of $\mathbf{S}\neq$ $max(\underset{col}{\mathrm{min})}$ of $\mathbf{S}$. As it can be seen in Table II, the $min(\underset{row}{\mathrm{max})}=7.71$ and $max(\underset{col}{\mathrm{min})}=0$, also in Table III $min(\underset{row}{\mathrm{max})}=0.76$ and $max(\underset{col}{\mathrm{min})}=0$. Since none of these two scenarios have a pure strategy, the mixed strategy of these two games have to be calculated. Based on the Tables II and III, and using Eq.~\ref{eq12}, the following optimization problems have to be solved to calculate the defenders' mixed strategies for the state estimator and the MLP, respectively. By solving Eq.~\ref{eq18}, $\tilde{q}$ can be calculated for the defender when the BDD of the defender is the state estimator. The mixed strategy of the defender when the state estimator is being used as the BDD can be calculated using Eq.~\ref{eq19}. 
\vspace{-5pt}
\begin{equation}
\begin{aligned}
&\underset{\tilde{q}}{\mathrm{max}}\;\tilde{q}^{'}\;1_{r}\;\;\;\;\;\;\;\;\;\;\;\;\;\;\;\;\;\;\;\;\;\;\;\;\;\;\;\;\;\;\;\;\;\;\;\;\;\;\;\;\;\;\;\;\; \\
&s.t.
\begin{cases}
2.79\tilde{q}_{2}\;+2.79\tilde{q}_{3}\;+4.45\tilde{q}_{4}\;+4.45\tilde{q}_{5}\;+7.71\tilde{q}_{6}\;\leq\;1\\
6.03\tilde{q}_{1}\;+6.03\tilde{q}_{3}\;+4.45\tilde{q}_{4}\;+10.44\tilde{q}_{5}\;+4.45\tilde{q}_{6}\;\leq\;1\\
5.83\tilde{q}_{1}\;+5.83\tilde{q}_{2}\;+10.18\tilde{q}_{4}\;+4.45\tilde{q}_{5}\;+4.45\tilde{q}_{6}\;\leq\;1\\
6.03\tilde{q}_{1}\;+2.79\tilde{q}_{2}\;+8.6\tilde{q}_{3}\;+6.03\tilde{q}_{5}\;+2.79\tilde{q}_{6}\;\leq\;1\\
5.83\tilde{q}_{1}\;+8.34\tilde{q}_{2}\;+2.79\tilde{q}_{3}\;+5.83\tilde{q}_{4}\;+2.79\tilde{q}_{6}\;\leq\;1\\
11.82\tilde{q}_{1}\;+5.83\tilde{q}_{2}\;+6.03\tilde{q}_{3}\;+5.83\tilde{q}_{4}\;+6.03\tilde{q}_{5}\;\leq\;1\\
\tilde{q}\;\geq\; 0.
\end{cases}
\end{aligned}
 \label{eq18}
\end{equation}
\begin{equation}
q=\tilde{q}\;h_{1}(q)=\tilde{q}(\tilde{q}1_{r})^{-1}
 \label{eq19}
\end{equation} 
\vspace{-0.25pt}
The mixed strategy of the defender when MLP is being used as the BDD can be calculated using Eq.~\ref{eq20}. Upon solving Eq.~\ref{eq20}, the mixed strategy of the defender when MLP is being used as the BDD can be calculated using Eq.~\ref{eq19}. Figs.~\ref{Fig4} \& ~\ref{Fig5} demonstrate the mixed strategies of these two games. As it can be seen in Fig. 4, the attacker performs 57\% of its attacks on $z4z5$, and this transmission line is defended by the defender in $21\%$ of the times. To find the mixed strategy of the attacker, Eq.~\ref{eq15} has to be solved using the values in Tables II and III respectively.
\begin{equation}
\begin{aligned}
&\underset{\tilde{q}}{\mathrm{max}}\;\tilde{q}^{'}\;1_{r}\;\;\;\;\;\;\;\;\;\;\;\;\;\;\;\;\;\;\;\;\;\;\;\;\;\;\;\;\;\;\;\;\;\;\;\;\;\;\;\;\;\;\;\;\; \\
&s.t.
\begin{cases}
0.86\tilde{q}_{2}\;+0.86\tilde{q}_{3}\;+0.28\tilde{q}_{4}\;+0.28\tilde{q}_{5}\;+0.77\tilde{q}_{6}\;\leq\;1\\
0.77\tilde{q}_{1}\;+0.77\tilde{q}_{3}\;+0.28\tilde{q}_{4}\;+0.78\tilde{q}_{5}\;+0.28\tilde{q}_{6}\;\leq\;1\\
1.5\tilde{q}_{1}\;+1.5\tilde{q}_{2}\;+1.28\tilde{q}_{4}\;+0.28\tilde{q}_{5}\;+0.28\tilde{q}_{6}\;\leq\;1\\
0.77\tilde{q}_{1}\;+0.86\tilde{q}_{2}\;+1.83\tilde{q}_{3}\;+0.77\tilde{q}_{5}\;+0.86\tilde{q}_{6}\;\leq\;1\\
1.5\tilde{q}_{1}\;+1.62\tilde{q}_{2}\;+0.86\tilde{q}_{3}\;+1.5\tilde{q}_{4}\;+0.86\tilde{q}_{6}\;\leq\;1\\
2.33\tilde{q}_{1}\;+1.5\tilde{q}_{2}\;+0.77\tilde{q}_{3}\;+1.5\tilde{q}_{4}\;+0.77\tilde{q}_{5}\;\leq\;1\\
\tilde{q}\;\geq\; 0.
\end{cases}
\end{aligned}
 \label{eq20}
\end{equation}

The main intuition that can be observed from Figs.~\ref{Fig4} \& ~\ref{Fig5} is that the probability of the attack is higher on the measurements that are less likely to be protected by the defender. For each set of measurement,s there is a corresponding value in $G$. The higher values of $G$ correspond to the measurements that have a greater effect on $\Delta\hat{\textbf{P}_{ij}}$. The adversary finds the measurements with greater value in $G$ and less probability of detection to compromise.   
The other observation from Figs.~\ref{Fig4} \& ~\ref{Fig5} is that, while the state estimator is being used as the BDD there is a larger number of measurements that can be compromised. As it can be seen, while the BDD is an MLP there are 3 sets of measurements that have a high probability to be compromised. However, as we use the state estimator as the BDD there are 4 sets of measurements which are likely to be compromised. This intuition means that using the MLP as the BDD not only decreases the gain that the adversary can achieve but also limits the set of possible actions that the adversary can choose to perform attacks.

This behavior can be explained using Tables II \& III. As it can be seen in Table II, in the scenarios that the adversary can compromise two measurements at the same without being defended it always gains more utility. As an example, while the adversary compromises $z1z4$ and the defender protects $z5z10$ the utility of the attacker is equal to 7.71MW. This behavior can be observed in other columns of Table II as well. However, this behavior cannot be seen in Table III.  In the scenario where the adversary attacks $z1z4$ and the defender protects $z5z10,$ the utility that the adversary gains are equal to 0.77MW which is not greater than all of the utilities that the adversary can gain in the first column of Table I. This means that while we use the MLP as the BDD there is no guarantee for the adversary to gain higher utilities if it can successfully compromise two meters simultaneously.  

\section{conclusion}
In this paper, an approach has been proposed which for the first time combines the MLPs with the game theory to study the false data injection attacks in the smart grids. This paper proposes to use the MLPs instead of conventional state estimators as the BDD of the defender in the game-theoretic formulation of the smart grid security. We have shown that if the MLPs are used then the adversaries can gain much less utility than when the state estimators are used. We have also studied the adversary informed and adversary uninformed scenarios in the game-theoretic formulation of this problem, and we have shown that depending on the type of the defender's BDD in each scenario different behavior can be observed. We also showed that this game does not have a pure strategy, and the mixed strategy has been calculated.

\bibliographystyle{ACM-Reference-Format}
\bibliography{sample-base}


 \end{document}